\documentclass[10pt,letterpaper]{article}

\usepackage{cogsci}
\usepackage{pslatex}
\usepackage{apacite}
\usepackage{float}
\usepackage{graphicx}
\usepackage{multirow}
\usepackage{todonotes}
\usepackage{booktabs}
\usepackage{xcolor}

\title{A Linguistic Analysis of Spontaneous Thoughts: Investigating Experiences of Deja Vu, Unexpected Thoughts, and Involuntary Autobiographical Memories}

\title{A Linguistic Analysis of Spontaneous Thoughts: Investigating Experiences of Déjà Vu, Unexpected Thoughts, and Involuntary Autobiographical Memories}

\author{
Videep Venkatesha\textsuperscript{1}, 
Mary Cati Poulos\textsuperscript{2}, 
Christopher Steadman\textsuperscript{2}, 
Caitlin Mills\textsuperscript{2}, 
Anne M. Cleary\textsuperscript{3}, 
Nathaniel Blanchard\textsuperscript{1} \\
\textsuperscript{1}Department of Computer Science, Colorado State University, Fort Collins, USA \\
\textsuperscript{2}Department of Educational Psychology, University of Minnesota, Twin Cities, USA \\
\textsuperscript{3}Department of Psychology, Colorado State University, Fort Collins, USA
}

\begin{document}

\maketitle
\begin{abstract}
The onset of spontaneous thoughts are reflective of dynamic interactions between cognition, emotion, and attention. Typically, these experiences are studied through subjective appraisals that focus on their triggers, phenomenology, and emotional salience. In this work, we use linguistic signatures to investigate Déjà Vu, Involuntary Autobiographical Memories, and Unexpected Thoughts. Specifically, we analyze the inherent characteristics of the linguistic patterns in participant generated descriptions of these thought types. We show how, by positioning language as a window into spontaneous cognition, existing theories on these attentional states can be updated and reaffirmed. Our findings align with prior research, reinforcing that Déjà Vu is a metacognitive experience characterized by abstract and spatial language, Involuntary Autobiographical Memories are rich in personal and emotionally significant detail, and Unexpected Thoughts are marked by unpredictability and cognitive disruption. This work is demonstrative of languages' potential to reveal deeper insights into how internal spontaneous cognitive states manifest through expression.



\textbf{Keywords:} 
Spontaneous Thoughts; Machine Learning; Natural Language Processing, Phenomenology
\end{abstract}

\section{Introduction}

Spontaneous thoughts are mental states or sequences of mental states that occur often in our daily lives. They are a cornerstone of human cognition, occupying as much as 30–50\% of our waking lives \cite{killingsworth2010wandering}. They are thought to arise due to an absence of strong constraints such as deliberate focus, external demands, or specific tasks that guide our attention or thought  \cite{christoff2016mind}. For example, you walk into a cafe for the first time, and you suddenly think to yourself, “I feel like I’ve been here before, but I can’t pinpoint why”. This sudden, compelling sensation that a situation has been experienced before despite evidence to the contrary is known as déjà vu, a form of spontaneous thought \cite{cleary2012familiarity}.


Historically, the study of spontaneous thoughts has relied on participants’ self-reports of such experiences as well as their corresponding appraisal ratings \cite{berntsen2021involuntary, cleary2012familiarity}. For example, Involuntary Autobiographical Memories (IAM), the recollection of personal events triggered by environmental cues, are often recorded in diary studies where participants log memory occurrences and rate them on various dimensions \cite{berntsen1996involuntary}, while unexpected thoughts (UT), thoughts that feel surprising in timing and content and offer new insights or novel perspectives \cite{poulos2023investigating,berntsen1996involuntary}, have been assessed through structured prompts that assess prior experiences with such phenomena \cite{poulos2023investigating, steadman2024involuntary}. 

These approaches have allowed us to describe the phenomenological patterns and potential triggers of such experiences by having participants reflect on dimensions such as emotional valence and their possible cues. At the same time, utilizing appraisal ratings alone neglects the linguistic dimension of the descriptions provided by participants, or the ways in which these experiences are described. In this work, we address this gap by leveraging the linguistic information provided by participants in a study that has been previously published \citeA{steadman2024involuntary} to examine three distinct spontaneous thought types – déjà vu, IAMs, and UTs. 

We expand on previous work by~\citeA{steadman2024involuntary}, which showed several key differences between the three spontaneous thought types mentioned above, among older and younger adults. Their findings indicated that older adults tended to rate their involuntary thoughts as more spontaneous, absorbing, and unrelated to their current task, while younger adults described déjà vu as more positive compared to IAMs and UTs. To show these differences, \citeA{steadman2024involuntary} relied on principal components analysis and sentiment analysis from participants’ descriptions and appraisals of these past experiences, along with machine learning classification models to predict participants’ age and the type of involuntary thought. We extend their study by investigating the linguistic characteristics of spontaneous thoughts, leveraging natural language processing techniques to analyze how participants describe their experiences. 

We chose to examine the content of participants' provided thought descriptions as, language provides a window into how individuals encode, frame, and express these mental experiences, reflecting both cognitive and emotional nuances.  
The idea that language forms the bedrock of cognition has been widely discussed, with researchers arguing that linguistic structures provide a framework for conceptualization and reasoning \cite{Levinson2003}.  Furthermore, language has been shown to be central to emotional experience, playing a crucial role in the construction and expression of affective states \cite{Lindquist2015} and in the communication of emotions in social contexts \cite{Reisenzein2012}. Thus, how individuals articulate and frame their thoughts through language can reveal subtle patterns within spontaneous thought types and may provide insight into potential similarities and differences among different forms of spontaneous thought. 

We hypothesized that linguistic features would meaningfully distinguish Déjà Vu, Involuntary Autobiographical Memories, and Unexpected Thoughts, echoing prior appraisal-based findings. Specifically, we expected language to reflect known theoretical distinctions such as DV’s abstract and generally positive tone \cite{cleary2021deja, mcneely2023piquing}, IAMs’ vivid personal detail, and UTs’ surprising and often negative content \cite{poulos2023investigating}, while remaining open to emergent patterns through exploratory analysis.

We conduct our analysis using two language representations: Term Frequency-Inverse Document Frequency, a common method for highlighting important words in text, and contextual embeddings from BERT, which capture meaning based on context within a sentence. These methods allow us to assess both surface-level word usage and deeper semantic patterns within participants’ descriptions. Additionally, we employ logistic regression coefficient analysis to identify words most predictive of each thought type, shedding light on the specific linguistic markers associated with spontaneous thought experiences. Beyond classification, we also analyze the emotional content of these descriptions using pre-trained models, capturing the affective landscape of each thought type.

Our findings reveal that spontaneous thoughts exhibit distinct linguistic signatures. Déjà vu experiences are more likely to contain abstract and spatial terms, whereas IAMs are characterized by vivid, autobiographical details. UTs, on the other hand, often include markers of surprise, intensity, and intrusiveness. Classification models achieve over 70\% accuracy, demonstrating that language alone provides meaningful differentiation between these thought types. Furthermore, emotional analysis highlights key differences in affective tone, with déjà vu descriptions showing a greater proportion of neutral and positive emotions, while IAMs and unexpected thoughts tend to contain more negative emotional content.  These findings align with existing research on the phenomenology of spontaneous thoughts, which suggests that IAMs are more emotionally intense and personally significant \cite{berntsen2002emotionally}, while déjà vu is often described as a neutral or even slightly positive metacognitive experience \cite{cleary2021deja,mcneely2023piquing}. Similarly, prior work on UTs highlights their intrusive and unpredictable nature \cite{poulos2023investigating}.

\section{Methods}

\subsection{Dataset}


Data from 314 participants were analyzed to examine three types of involuntary thought experiences: Déjà vu, Involuntary Autobiographical Memory (IAM), and Unexpected Thought (UT). Participants were instructed to recall and type out one UT, one IAM, and one instance of déjà vu, for a total of three thoughts. These descriptions formed the basis of the text classification models and linguistic analyses conducted in our study.
The dataset consisted of detailed textual descriptions of involuntary thoughts, with additional metadata capturing appraisal dimensions such as spontaneity, relatedness to the task, and emotional intensity.  The prompts are detailed below. Additional details on the dataset are available in \citeA{steadman2024involuntary}.  
\begin{itemize}
    \item Déjà vu: Think back to a time when you felt strangely like a situation was a re-experience of something that you've experienced before, but could not pinpoint why.
    \item IAM: Think back to a time when you had a specific memory involuntarily pop into your head.
    \item UT: Think back to a time when you had an unexpected thought involuntarily pop into your head.
\end{itemize} 
 All of the full recall data is also available at \mbox{\texttt{https://osf.io/ge3f8/?view\_only=}}
 \mbox{\texttt{2826a7aa65a34e77a4c22b77a159194a}}


\subsection{Language Representation}

Two types of language representations were used: Term Frequency-Inverse Document Frequency (TF-IDF) \cite{manning2008introduction} and BERT-base-uncased \cite{devlin2018bert}. These approaches provide both basic and richer contextual embeddings, allowing us to evaluate language patterns at different levels of complexity.

\subsubsection{TF-IDF Representation}
The TF-IDF representation is a traditional method that transforms text into numerical values based on word frequency, allowing for a straightforward yet informative depiction of a text’s vocabulary. TF-IDF emphasizes words that are unique to specific documents within a larger collection, aiming to highlight distinctive language features. The “Term Frequency” (TF) component measures how often a word appears in a document, while the “Inverse Document Frequency” (IDF) component downweights words that appear frequently across many documents. By combining these factors, TF-IDF assigns a higher weight to words that are important within individual documents but rare across the entire set of documents, helping to distinguish each thought type based on characteristic vocabulary.

For instance, if the word “memory” appears frequently in descriptions of IAMs but less so in descriptions of déjà vu or UTs, TF-IDF would assign it a relatively high weight in IAM-related texts. This method provides a surface-level representation that is easy to interpret and highlights distinctive terms for each thought type without relying on the broader context in which they appear.
\subsubsection{BERT-base-uncased Representation}
TF-IDF captures the frequency and uniqueness of words but it lacks the ability to understand word meanings in context. To address this limitation, the BERT-base-uncased, a transformer-based model was used, that can generate richer, context-sensitive embeddings for each thought type. BERT (Bidirectional Encoder Representations from Transformers) is a pre-trained model designed to capture nuanced language patterns by examining the context of each word. BERT considers each word in relation to its surrounding words, generating embeddings that reflect both the individual word meanings and the overall sentence context, unlike traditional word representations, which treat words independently \cite{devlin2018bert}.


\subsection{Text Classification}
Text classification machine learning models were trained to automatically classify participant thought types. Model performance was assessed using accuracy, Cohen’s Kappa \cite{mchugh2012interrater}, and the F1 score.
Accuracy represents the proportion of correct predictions out of the total number of predictions. In contrast, Cohen's Kappa evaluates the agreement between predicted and actual values while accounting for agreement occurring by chance. 
Kappa values range from -1 (complete disagreement) to 1 (perfect agreement), with a value of 0 indicating that the model's performance is equivalent to random guessing. Finally, F1 score is the harmonic mean of precision and recall, where precision is the ratio of total positive to total predicted positive, and recall is the ratio of total positive to total actual positive. 

\subsubsection{Text Preprocessing}
\vspace{-1mm}
We applied two sets of analyses to prepare the descriptions of involuntary thoughts for text classification. Text preprocessing refers to standard methods for cleaning and preparing text for computational analysis. We conducted one analysis without any preprocessing and one with preprocessing. The preprocessing aimed to reduce potential biases that could arise from participants explicitly stating the type of thought (e.g., stating ``I experienced déjà vu when…”). This is a well established technique used in Natural Language Processing \cite{sarica2021stopwords}. We thus applied the following text preprocessing --- a) Stop Word Removal: Common stop words (e.g., `the', `is' and were removed and b) Keyword Removal: Words explicitly indicating the type of thought (e.g., `déjà vu', `unexpected', `involuntary', `popped') were removed. For example, the sentence ``I once had an unexpected thought to leave the UK and emigrate to New Zealand against the wishes of my family" becomes ``leave UK emigrate New Zealand against wishes family".


\subsubsection{Model and Hyperparameter selection}
Several machine learning models such as logistic regression, decision tree, random forest, fasttext, and support vector machine (SVM) were used for the TF-IDF representation. Hyperopt was used to fine-tune hyperparameters. This library utilizes a probabilistic model to systematically search for hyperparameters that optimize the model's performance metric. In our case, we sought to maximize the weighted average F1 score over a 1000 trials. We selected the model with median F1 performance in order to analyze the general trends, rather than risking drawing conclusions from an outlier model. We also note that the model search returns multiple models with the same output since slight changes in the hyperparameters might not cause a significant change in the model performance. The duplicate models were removed and the results of the median model are presented in the result section. For the richer BERT representations, we employed a grid search to converge on a set of hyperparameters. Regardless of language representation, each model was trained and evaluated using leave-one-participant-out cross-validation

\subsubsection{Coefficient Analysis for Logistic Regression}
A logistic regression classifier trained on the TF-IDF representation of participant descriptions was used for the analysis of the linguistic features. Logistic regression allows for the examination of feature coefficients, which provide insights into how individual words contribute to the classification of thought types. Each feature (word) in the dataset is assigned a weight, indicating its positive or negative association with a specific thought type.
To identify these associations, we extracted the top 10 positively and negatively weighted coefficients for each thought type. These coefficients represent the linguistic features most predictive of each class, providing a foundation for understanding the distinctive language patterns associated with déjà vu, IAMs, and UTs.

\subsection{Emotion Analysis}

An emotion analysis was conducted to further investigate the inherent nature of these internal states. Specifically, we examined the specific emotions embedded in the thought descriptions. For this purpose, a pre-trained transformer model trained on 58k tweets was used. The text was not preprocessed in this analysis to maintain consistency with the original dataset the model was trained on \cite{saravia2018carer}. The model was used to classify emotions into categories: joy, sadness, fear, surprise, anger, or love.  

We added an additional layer of analysis, given that such models are often biased toward the data they are trained on, and recognizing that the domain of tweets might not fully translate to involuntary thought experiences. We utilized Llama 3.1 8-B \cite{grattafiori2024llama3}, a state-of-the-art large language model, to classify each thought description into the emotion categories motivated by Ekman’s Theory of Emotions \cite{ekman1992there}. To accommodate the fact that some thoughts might not contain significant emotional content, we also included a 'neutral' category. This ensured a more nuanced understanding of the emotional landscape, allowing us to account for instances where the emotional content was either subtle or absent.

The combination of these approaches aims to minimize the potential biases and provides a more robust analysis of the emotional characteristics present in each thought type. By leveraging both a pre-trained model and a flexible prompting approach with Llama, we aimed to capture a broad spectrum of emotions conveyed in the participants' reports.

\section{Results}
We begin by presenting the results of the text classification models. We then present the results of the coefficient analyses to reveal the distinct words that allowed the models to distinguish one thought type from another. Finally, we present the emotion analyses, which examined the distribution of emotions across the different thought types using both the BERT model and the Llama 3.1 prompt-based approach.  
\subsection{Text Classification} 
The results for both TF-IDF and BERT representations are summarized in Table \ref{tab:metrics-compact}, with Table \ref{tab:confusion-matrix} providing the confusion matrices across all conditions. The matrices illustrate the number of correct and incorrect predictions made by each model for each thought type, allowing us to better understand the performance of each approach.


\begin{table}[ht]
    \centering
    \caption{Performance metrics across processing levels and language representation methods}
    \label{tab:metrics-compact}
    \vskip 0.12in
    \begin{tabular}{lcc|cc}
        \toprule
        & \multicolumn{2}{c}{\textbf{TF-IDF}} & \multicolumn{2}{c}{\textbf{BERT}} \\
        \cmidrule(lr){2-3} \cmidrule(lr){4-5}
        \textbf{Metric} & \textbf{Raw} & \textbf{Pre-proc.} & \textbf{Raw} & \textbf{Pre-proc.} \\
        \midrule
        Accuracy & 0.80 & 0.70 & 0.86 & 0.73 \\
        F1 & 0.81 & 0.70 & 0.86 & 0.73 \\
        Kappa & 0.71 & 0.60 & 0.80 & 0.58 \\
        \bottomrule
    \end{tabular}
\end{table}


As can be seen from Table \ref{tab:metrics-compact}, each text classification model was able to accurately classify each thought type with at least 70\% accuracy, suggesting that the language used to describe these involuntary thought experiences is quite separable while still suggesting overlapping features amongst them. The raw descriptions with BERT representations performed the best, achieving 86\% accuracy. However, it is important to note that there is a significant drop in performance for both the TF-IDF and BERT models after pre-processing, likely because many of the provided thought descriptions contained keywords (e.g., memory) that improved overall model performance. 
Notably, the models utilizing contextual embeddings (BERT) did not significantly outperform traditional approaches like TF-IDF in accuracy and F1 scores. This suggests that while contextual embeddings provide added depth of understanding, the improvement over traditional methods like TF-IDF is incremental, which underscores the complexity of distinguishing between thought types. 


\begin{table}[ht]
\centering
\caption{Confusion Matrices for Déjà Vu (DV), Involuntary Autobiographical Memories (IAM), and Unexpected Thoughts (UT) using Term Frequency-Inverse Document Frequency (TF-IDF) and Bidirectional Encoder Representations from Transformers (BERT) models}
\label{tab:confusion-matrix}
\vskip 0.12in
\renewcommand{\arraystretch}{1.2}
\setlength{\tabcolsep}{5pt}

\begin{tabular}{ccccc}
\hline
\multirow{2}{*}{\textbf{Model}} & \multirow{2}{*}{\textbf{Actual}} &
  \multicolumn{3}{c}{\textbf{Predicted}} \\ 
\cline{3-5}
  &  & \textbf{Déjà Vu} & \textbf{IAM} & \textbf{UT} \\
\hline

\multirow{3}{*}{\shortstack{\textbf{Raw}\\\textbf{TF-IDF}}}
  & \textbf{Déjà Vu} & 258 & 37  & 19  \\
  & \textbf{IAM}     & 18  & 253 & 43  \\
  & \textbf{UT}      & 16  & 16  & 252 \\
\hline

\multirow{3}{*}{\shortstack{\textbf{Preprocessed}\\\textbf{TF-IDF}}}
  & \textbf{Déjà Vu} & 237 & 41  & 36  \\
  & \textbf{IAM}     & 21  & 213 & 80  \\
  & \textbf{UT}      & 19  & 86  & 209 \\
\hline

\multirow{3}{*}{\shortstack{\textbf{Raw}\\\textbf{BERT}}}
  & \textbf{Déjà Vu} & 280 & 23  & 11  \\
  & \textbf{IAM}     & 10  & 273 & 31  \\
  & \textbf{UT}      & 16  & 33  & 235 \\
\hline

\multirow{3}{*}{\shortstack{\textbf{Preprocessed}\\\textbf{BERT}}}
  & \textbf{Déjà Vu} & 253 & 31  & 30  \\
  & \textbf{IAM}     & 30  & 228 & 36  \\
  & \textbf{UT}      & 27  & 84  & 203 \\
\hline

\end{tabular}
\end{table}


Table \ref{tab:confusion-matrix} illustrates an interesting pattern in the data. Déjà vu exhibited greater separability (i.e., was less confused with other thought types) compared to Involuntary Autobiographical Memories (IAM) and Unexpected Thoughts (UT), which were consistently more likely to be confused with each other across all classification models. This suggests that IAMs and UTs may be more phenomenologically similar to one another, with déjà vu occupying a separate space in the phenomenology spectrum, a proposal consistent with previous findings \cite{poulos2023investigating, steadman2024involuntary}
\subsection{Coefficient Analysis}

We utilized a logistic regression classifier that was trained on the TF-IDF representation to further investigate the source of confusion amongst the thought types and, to identify the most and least influential tokens (i.e., positive and negative words) for predicting each thought type. These tokens thus reflect the subtle linguistic differences that distinguish déjà vu, IAM, and UT within the model. 
Déjà vu was characterized by positively weighted abstract and spatial terms (e.g., seemed, place, visited), while negatively weighted words were personal (e.g., mother, father, child), highlighting its contentless, metacognitive nature \citeA{cleary2023flips, mcneely2023piquing,neisser2023opening} 

IAMs showed the opposite trend, positively weighted words were personal (e.g., parents, girlfriend, mum), while negatively weighted words referenced knowledge or experience (e.g., knew, happened, strange), aligning with findings that IAMs are vivid and autobiographical \cite{berntsen1996involuntary, steadman2024involuntary}

UTs  were marked by positively weighted emotionally charged words (e.g., unexpectedly, death, urge, random), while also having a few terms that were personal and knowledge-based, similar to IAMs. This linguistic overlap likely explains why IAMs and UTs were more frequently confused with each other than with déjà vu \cite{poulos2023investigating, steadman2024involuntary}
\subsection{Emotion Analysis}
\begin{figure}[ht]
    \centering
    \includegraphics[width=0.85\linewidth]{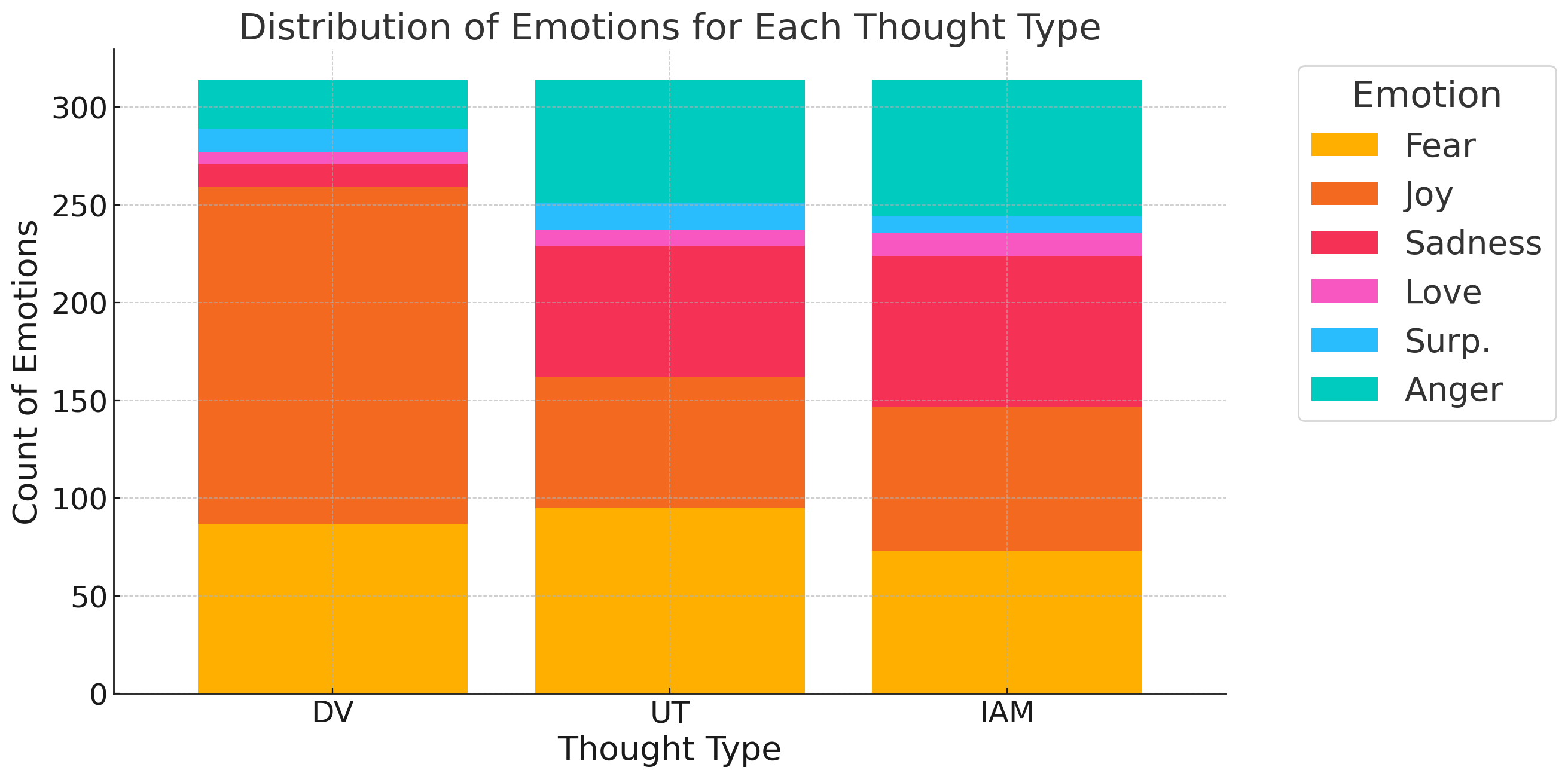}
    \caption{Distribution of emotions for each thought type using BERT. The model classified sentences into one of six emotion categories: Fear, Joy, Sadness, Disgust, Surprise, Anger.}
    \label{fig:emotion-distribution-bert}
\end{figure}

Finally, we analyzed the emotional content of the thought reports using two models --- a BERT based model trained on Twitter data and a Llama 3.1 zero shot emotion classification. Figure \ref{fig:emotion-distribution-bert} presents the results of the BERT-based model. The model classifies a sentence into Joy, Fear, Sadness, Love, Surprise, and Anger.  As can be seen from figure \ref{fig:emotion-distribution-bert}, the BERT-based model indicated that déjà vu reports contained more words related to joy and fewer words related to sadness and anger compared to UT and IAM. UTs and IAMs shared a greater overlap in emotional tone, with IAMs generally carrying fewer fear-related terms. 


\begin{figure}[ht]
    \centering
    \includegraphics[width=0.85\linewidth]{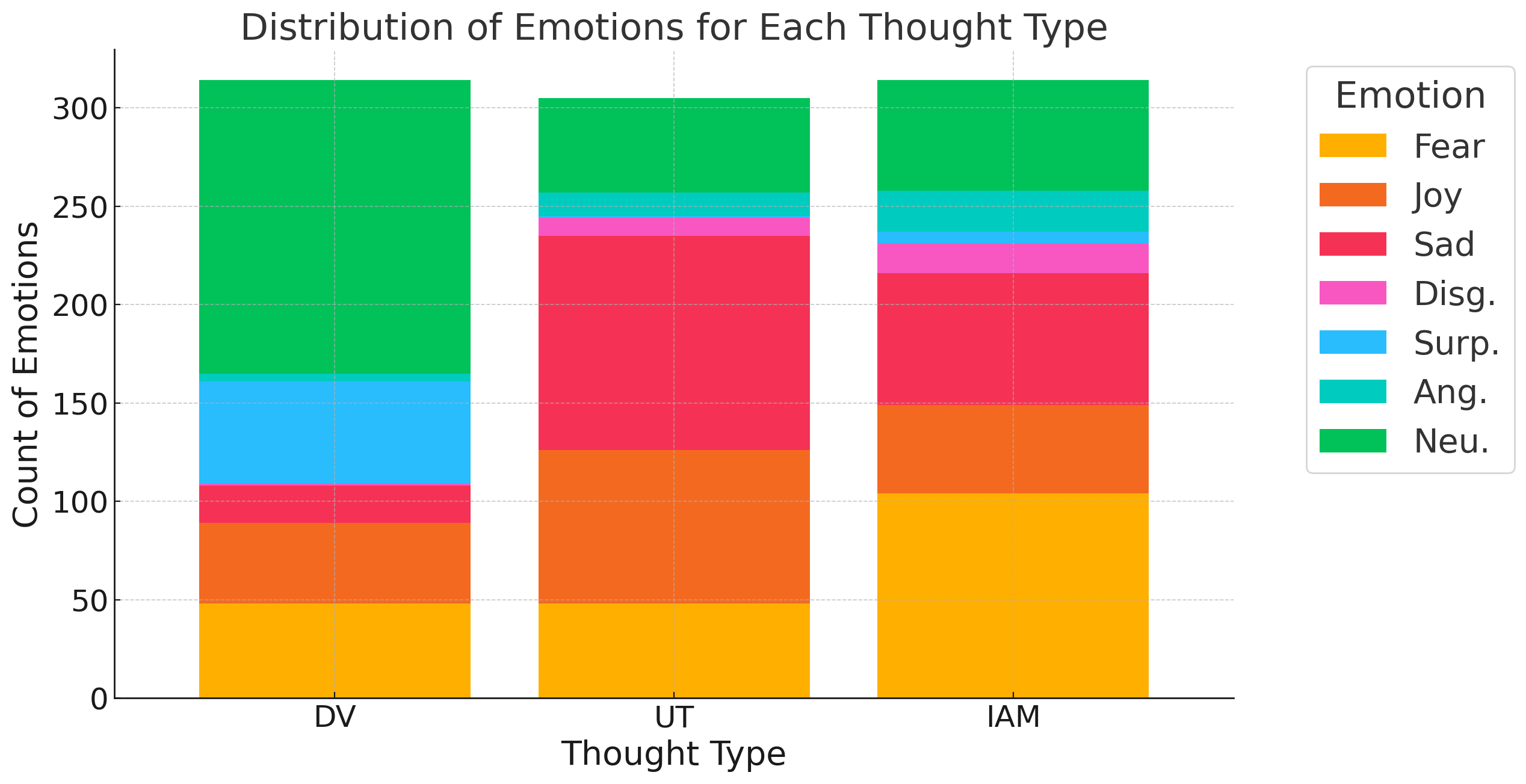}
    \caption{Distribution of emotions for each thought type using LLaMA 3.1. The model classified sentences into one of seven emotion categories: Fear, Joy, Sadness, Disgust, Surprise, Anger, and Neutral.}
    \label{fig:emotion-distribution-llama}
\end{figure}

We then prompted Llama 3.1 with 7 emotion categories where the intention was to extract the emotion from the content of the experience. The model classified each sentence into one of the emotions: Fear, Joy, Sadness, Disgust, Surprise, Anger, and Neutral.  The distribution of these emotions for each thought type is summarized in Figure \ref{fig:emotion-distribution-llama}. The Llama model provided a more nuanced understanding of the emotional landscape by including a `neutral' category, which helped to capture the subtler aspects of emotional content. Furthermore, the prompt was designed to extract the emotions of the content of the thought. The analysis revealed that déjà vu was characterized by a higher proportion of neutral and joy-related language, while IAM was associated with more fear-related terms, and UT exhibited more varied emotional content, including sadness and disgust.

\section{  Discussion }

Our findings revealed intriguing patterns that both align with and extend existing literature. Our results supported our initial hypothesis that language alone could meaningfully differentiate spontaneous thought types in ways consistent with prior appraisal-based methods. First, using our novel approach, which included leveraging richer language representations and selecting the model with median performance, we replicated known distinctions between thought types, with déjà vu showing greater separability and IAMs and UTs frequently confused—patterns \cite{steadman2024involuntary}. Patterns of misclassification between IAMs and UTs further emphasized the nuanced overlap between these two thought types. IAMs often contained specific and vivid details, leading to frequent confusion with UTs, particularly when emotionally significant content was present. In contrast, déjà vu descriptions, despite occasionally including concrete narrative elements, were largely distinct, further underscoring their abstract nature. These trends highlight how both linguistic content and emotional tone play a crucial role in distinguishing thought types.

To further explore how the models classified each thought type, we utilized a coefficient analysis, which revealed several intriguing patterns. For déjà vu, positively weighted terms included neutral or spatial descriptors such as ``place", ``walking", and ``seemed" again suggesting a sense of abstract familiarity and also consistent with research showing that déjà vu most commonly occurs with places \cite{cleary2021deja}. Conversely, negatively weighted terms, such as ``mother," ``father," and ``child," underscored the absence of strong personal or emotionally charged connections, reinforcing the idea that déjà vu may be characterized as a phenomenon that is contentless in nature \cite{cleary2023possible}. This aligns with theories of metamemory \cite{neisser2023opening}, which posit that the feeling of familiarity without specific content is a distinct attentional state with this study providing computational evidence for its abstract and spatially oriented nature.

In contrast, IAMs were characterized by positively weighted terms related to vivid, personal, and relational content, such as ``dinner," ``party," and ``father," reflecting their autobiographical essence \cite{berntsen1996involuntary}. Negative terms for IAM, including ``strange" and ``walking," highlighted the contrast with the concrete, detailed nature of these memories. For unexpected thoughts, positively weighted terms like ``random," ``intrusive," and ``unexpectedly" captured their sudden and surprising quality, while negative terms, such as ``specific" and ``visited," indicated that UTs are less tied to structured or historical details. 


The exploration of emotional content within the descriptions of spontaneous thoughts was motivated by prior research highlighting the integral role of emotions in cognitive phenomena such as memory retrieval, attention shifts, and mind-wandering.  For example, \citeA{berntsen1996involuntary} emphasized the emotional vibrancy of involuntary autobiographical memories, while~\citeA{poulos2023investigating} and \cite{steadman2024involuntary} identified the emotionally charged nature of unexpected thoughts. Similarly, the association of déjà vu with curiosity and positive emotional states \cite{mcneely2023piquing} provided a rationale for investigating emotional dimensions across all three thought types. Our findings revealed distinct emotional profiles.

Before delving into the specific results, it is important to note that the emotion analysis broadly supports the classification analyses by reinforcing the observed distinctions among the three thought types. Specifically, déjà vu, which was consistently classified as the most distinct thought type, also exhibited a unique emotional profile characterized by neutrality and occasional positivity. In contrast, the overlap in emotional profiles between UTs and IAMs --— both displaying predominantly negative emotional content—parallels their higher misclassification rates.

Déjà vu consistently exhibited more neutral or positive emotional content, aligning with its characterization as a phenomenon marked by a sense of familiarity without specific or emotionally charged content  \cite{cleary2023possible,mcneely2023piquing}. UTs demonstrated a predominance of negative emotional content, with higher levels of fear and sadness. These thoughts often emerge as startling or disturbing, with emotionally charged content that disrupts attention. The Llama model's classification of UTs included significant markers of `negative feelings,' mainly fear, underscoring their intrusive and distressing impact. Similarly, IAMs also exhibited predominantly negative emotions, but these were more frequently manifested as sadness rather than fear. This distinction emphasizes the autobiographical and emotionally reflective nature of IAMs.


This study is also motivated by the broader implications of classifying attentional states. Detecting attention shifts is a growing area of interest in applied domains such as education and adaptive learning systems \cite{blanchard2014automated, smith2018mind, castillon2022multimodal, venkatesha2024propositional}. For example, given the association between déjà vu and curiosity \cite{mcneely2023piquing}, identifying instances of déjà vu could signal teachable moments where learners are primed for exploration, while UT, akin to mind-wandering, might require intervention to refocus attention. By identifying not just when attention shifts occur but also the nature of the underlying thought, ML models could pave the way for personalized interventions that enhance engagement and learning outcomes.

Several limitations must be acknowledged in the current study. The use of natural language processing to identify linguistic patterns may inadvertently highlight the structure of language rather than the inherent nature of the thoughts themselves. However, despite this potential limitation, our findings align with existing theories on spontaneous cognition, suggesting that the linguistic patterns captured are reflective of underlying cognitive states rather than merely artifacts of language use. Additionally, the reliance on retrospective self-reports introduces the challenge of memory accuracy, as recollection may not fully capture the immediacy or authenticity of these thoughts in real time. This limitation underscores the difficulty of detecting and analyzing spontaneous cognitive states as they occur. The reliance on self-reported descriptions introduces subjectivity, as participants may differ in their ability to articulate their thoughts or may selectively report details. This limitation is compounded by linguistic and cultural variability; the study focused exclusively on English-speaking participants, which restricts the generalizability of findings to other languages and cultures. Another limitation lies in the potential overfitting of ML models to dataset-specific patterns, despite the use of cross-validation techniques and the use of the median model. 
We also acknowledge that language may not provide a fully unfiltered window into internal cognitive states, and that descriptions are inevitably shaped by participants' conceptualizations and familiarity with certain terms. Our preprocessing approach aimed to remove the most overt self-labels (e.g., `déjà vu', `unexpected thought') to reduce trivial classification, while preserving more naturalistic descriptors like `intrusive', which participants might use spontaneously rather than diagnostically. Future work could more systematically investigate the boundary between conscious self-ascription and underlying linguistic signals. Additionally, while we envision applications for improving educational and clinical interventions, we recognize that language-based models of spontaneous thought raise important ethical concerns, including the potential for misuse in domains such as targeted advertising or persuasion.

The present findings open several avenues for future research. First, the ability to distinguish spontaneous thought types using linguistic patterns suggests potential applications in clinical memory assessments, where spontaneous recollections could serve as non-invasive cognitive markers. Second, the alignment between linguistic signatures and cognitive theories motivates the development of computational models of spontaneous thought that incorporate language-based features, enabling more precise modeling of spontaneous retrieval and prediction errors in cognition.

\subsubsection{Conclusion}

This study demonstrates the viability of using language as a methodological lens to study involuntary thought experiences and internal cognitive states. While our findings reaffirm known differences between spontaneous thoughts, they do so through linguistic and computational means, offering converging evidence that language carries signatures aligned with existing psychological theories. Our approach holds particular promise for investigating less well-understood phenomena, where expectations are less defined. We propose that language-based analysis should be viewed as a complementary modality in the study of spontaneous cognition, capable of revealing both surface-level features and deeper patterns of meaning that might otherwise remain inaccessible.

\section{Acknowledgements}
This material is based in part upon work supported by the National Science Foundation (NSF) under
subcontracts to Colorado State University on award DRL 2019805 (Institute
for Student-AI Teaming), and by Other Transaction award HR00112490377 from the U.S. Defense Advanced Research Projects Agency (DARPA) Friction for Accountability in Conversational Transactions
(FACT) program. Approved for public release, distribution unlimited. Views expressed herein do not
reflect the policy or position of the National Science Foundation, the Department of Defense, or the U.S.
Government. All errors are the responsibility of the authors

\bibliographystyle{apacite}

\setlength{\bibleftmargin}{.125in}
\setlength{\bibindent}{-\bibleftmargin}

\bibliography{CogSci_Template}

\end{document}